\documentclass[conference]{IEEEtran}
\usepackage{xcolor}

\usepackage{caption}
\usepackage{subcaption}
\usepackage{graphicx}
\usepackage{tabularray}
\usepackage{multirow}
\usepackage{tabularx}
\usepackage{booktabs}
\usepackage{tablefootnote}
\usepackage{amsmath}
\usepackage[flushleft]{threeparttable}
\usepackage[symbol]{footmisc}

\usepackage{fancyhdr}
\fancypagestyle{firstpage}{
  \fancyhf{}
  \fancyhead[C]{To appear at the IEEE World Congress on Computational Intelligence (IEEE WCCI 2024)\\ at Yokohama, Japan, 30 June - 5 July 2024.}
}

\begin{document}
\title{ApproxDARTS: Differentiable Neural Architecture Search with Approximate Multipliers}

\author{
\IEEEauthorblockN{Michal Pinos}
\IEEEauthorblockA{\textit{Faculty of Information Technology}\\\textit{Brno University of Technology}\\ Brno, Czech Republic\\ipinos@fit.vutbr.cz}
\and
\IEEEauthorblockN{Lukas Sekanina}
\IEEEauthorblockA{\textit{Faculty of Information Technology}\\\textit{Brno University of Technology}\\ Brno, Czech Republic\\sekanina@fit.vutbr.cz}
\and
\IEEEauthorblockN{Vojtech Mrazek}
\IEEEauthorblockA{\textit{Faculty of Information Technology}\\\textit{Brno University of Technology}\\ Brno, Czech Republic\\mrazek@fit.vutbr.cz}

}

\maketitle

\thispagestyle{firstpage}
\begin{abstract}
Integrating the principles of approximate computing into the design of hardware-aware deep neural networks (DNN) has led to DNNs implementations showing good output quality and highly optimized hardware parameters such as low latency or inference energy. In this work, we present ApproxDARTS, a neural architecture search (NAS) method enabling the popular differentiable neural architecture search method called DARTS to exploit approximate multipliers and thus reduce the power consumption of generated neural networks. 
We showed on the CIFAR-10 data set that the ApproxDARTS is able to perform a complete architecture search within less than $10$ GPU hours and produce competitive convolutional neural networks (CNN) containing approximate multipliers in convolutional layers. For example, ApproxDARTS created a CNN showing an energy consumption reduction of (a) $53.84\%$ in the arithmetic operations of the inference phase compared to the CNN utilizing the native $32$-bit floating-point multipliers and (b) $5.97\%$ compared to the CNN utilizing the exact $8$-bit fixed-point multipliers, in both cases with a negligible accuracy drop. Moreover, the ApproxDARTS is $2.3\times$ faster than a similar but evolutionary algorithm-based method called EvoApproxNAS. 
\end{abstract}


%
\IEEEpeerreviewmaketitle

\section{Introduction}

Automated design of \emph{deep neural networks} (DNN), known as \emph{neural architecture search} (NAS)~\cite{NAS:ACMSurv:2021, zoph2017neural} enabled designers to reduce the design time and obtain human-competitive or even better neural networks. For example, image classifiers based on \emph{convolutional neural networks} (CNNs) created by NAS methods exhibit better trade-offs between the classification accuracy and inference latency than human-created CNNs under the so-called \emph{mobile setup}, in which the number of multiply and accumulate (MAC) operations per one inference is bounded by 600 million~\cite{Sze:book:2020, Benmeziane:ijcai2021}. This type of NAS assumes that a CNN is designed for a particular hardware, i.e., a mobile phone in the aforementioned example. 

Originally, NAS methods utilized computationally expensive black-box methods such as reinforcement learning and evolutionary algorithms to deliver DNN architecture~\cite{NAS:insightfrom1000papers}. Later, \emph{differentiable architecture search} (DARTS), exploiting a continuous relaxation of the discrete architecture search space and performing a joint optimization of the architecture and weights using a gradient descent method led to significantly shortening the DNN design time~\cite{NAS:insightfrom1000papers}.
And, very recently, a new research direction was established -- \emph{NAS with hardware co-optimization} -- in which DNN model and hardware configuration of a configurable hardware accelerator are co-designed in parallel. This approach, optimizing the computational cost of candidate DNNs by selecting a suitable pruning, bit precision, quantization, and hardware mapping directly within NAS, enabled additional improvement of the accuracy-energy-latency trade-offs~\cite{Benmeziane:ijcai2021, NASCosearchSurvey}.

This paper deals with introducing selected principles of \emph{approximate computing} to hardware-aware NAS methods. The idea of approximate computing is to find the best trade-off between the accuracy and hardware parameters provided by an approximate implementation~\cite{AxCinDNN:survey:2023}. It exploits the fact that DNN implementations are highly error-resilient, so inexact components (such as approximate multipliers) can be used to implement some of their operations with a minimum impact on accuracy. Note that the approximation is usually performed \emph{a posteriory}, i.e., after the DNN architecture is fixed. Contrasted, for example, EvoApproxNAS~\cite{Pinos:EvoApproxNAS} utilized an evolutionary algorithm to design energy-efficient CNN-based image classifiers by evolving CNN architecture and concurrently assigning approximate multipliers to multiplication operations in convolutional layers. However, introducing approximate multipliers into NAS slowed the entire design process because approximate multipliers must be expensively emulated (usually using look-up tables) compared to the highly optimized arithmetic operations directly available on GPUs. 

This paper aims to introduce approximate multipliers to DARTS to obtain hardware-aware CNNs and reduce the design time compared to EvoApproxNAS. For the first time, we show that it is possible, even without introducing any significant modifications to the original version of DARTS.  
The proposed \emph{DARTS with approximate multipliers} (ApproxDARTS) is evaluated on CIFAR-10 benchmark, showing the reduced design time, improved accuracy and comparable trade-offs compared to CNNs produced by EvoApproxNAS.

The rest of the paper is organized as follows. Relevant work in the areas of NAS, hardware-aware NAS, and approximate computing is presented in Section~\ref{sec:soa}. In Section~\ref{sec:proposed}, we propose ApproxDARTS, a NAS method based on DARTS in which approximate multipliers can directly be employed. Section~\ref{sec:setup} summarizes the experimental setup used during the evaluation. Results are reported in Section~\ref{sec:results}. Conclusions are derived in Section~\ref{sec:conclusions}.

\section{Related Work}
\label{sec:soa}

This section surveys relevant research. As the paper deals with CNN-based image classifiers, we will only consider CNNs in the following sections.

\subsection{Neural Architecture Search}
\begin{figure}
    \centering
    \includegraphics[width=0.7\linewidth]{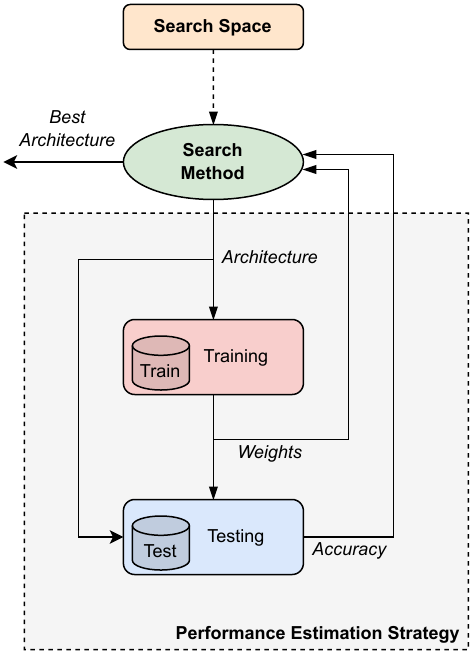}
    \caption{Three dimensions of NAS, i.e. \textit{Search Space} defining a space of all architectures that can be explored, \textit{Search Method}, implementing the algorithm for the architecture space exploration, and \textit{Performance Estimation Strategy}, responsible for the evaluation of  candidate architectures.}
    \label{fig:NAS}
\end{figure}

The origins of NAS date back to the 1980s~\cite{NAS:1989}; however, its popularity grew exponentially just recently~\cite{zoph2017neural}, with thousands of NAS papers being published since then~\cite{NAS:insightfrom1000papers}.
Throughout the years, NAS achieved some remarkable results in various tasks and in some cases, outperformed the best human-created CNN architectures~\cite{zoph2017neural}.
A diagram of NAS is shown in Figure~\ref{fig:NAS}.
Several different approaches for NAS have been explored, e.g reinforcement learning~\cite{ENAS,zoph2017neural}, Bayesian optimization~\cite{nguyen2021optimal} or evolutionary techniques~\cite{AmoebaNet:2019, Pinos:EvoApproxNAS, Suganuma:GECCO2017}. However, these techniques fall into the black-box optimization problem category and require substantial computational resources (e.g., $22\ 400$ GPU days in the NAS-RL~\cite{zoph2017neural}), as they rely on a large number of evaluations of candidate architectures.

In order to avoid the need to train and evaluate every candidate architecture from scratch, \emph{one-shot} techniques were introduced. Instead of individually training all architectures in the search space, one-shot techniques implicitly train all architectures via a single \emph{super-network} (\emph{supernet}), which contains all architectures from the search space as its \emph{sub-networks} (\emph{subnets}). After the supernet is trained, all architectures from the search space can be evaluated without the need for training, as the weights are simply inherited from the corresponding subnet within the supernet. Subsequently, one-shot methods are usually accompanied by a black-box method to search over the discrete architecture search space of the supernet to find the best-performing subnet (architecture).

A novel approach to one-shot techniques was introduced by DARTS~\cite{DARTS}, which uses a continuous relaxation of the discrete architecture search space and enables a joint optimization of the architecture and supernet weights using gradient descent. Due to its simplicity, small computational resource requirements, and competitive performance, DARTS attracted a lot of attention in the AutoML community and inspired a plethora of follow-up work~\cite{SGAS, XNAS, DSNAS, PCDARTS, PDARTS}. DARTS will be discussed in Section~\ref{sec:darts}.

\subsection{Hardware-Aware NAS}

The conventional NAS proved its efficiency by the ability to design state-of-the-art models; however, these models are usually composed of tens or hundreds of millions of parameters and require billions of MAC operations. These models are, therefore, too complex to be deployed in resource-constrained devices (e.g., on mobile devices, IoT gadgets, and embedded systems). For this purpose, a new type of NAS, called hardware-aware NAS (HW-NAS), was introduced.

HW-NAS differs from conventional NAS techniques in that it considers classification accuracy and other hardware properties of the implementation (e.g., energy consumption, latency, and MAC operations) during the search for the optimal model. The search algorithm is, in principle, a multi-objective optimization method. 
Various methods have been utilized, including a two-stage optimization (after reaching an acceptable quality according to one criterion, the second criterion can be optimized, as used in, e.g., OFA~\cite{Cai2020Once-for-All:}), constrained optimization (i.e., no candidate CNN is accepted that violates a given constraint, e.g., in MNASNet~\cite{MNASNet}, ProxylessNAS~\cite{ProxylessNAS}), scalarization (e.g., the objective are aggregated using the weighted sum), and truly-multiobjective optimizers constructing a Pareto front such as NSGA-II (NSGANet~\cite{NSGANet}).
If parameters of the hardware accelerator (such as the number of processing elements, cache setup, bit widths, CNN-to-HW mapping, etc.) are co-optimized, a new search space is, in addition to the space of CNN architectures, introduced. Then, the search algorithm searches for the best CNN architecture and hardware configuration in parallel. Such an approach can reach better implementations than independent optimizations of CNN architecture and hardware~\cite{NASCosearchSurvey}.

\subsection{Approximate Computing in CNNs}

Hardware implementations of CNNs utilize principles of approximate computing in which it is acceptable to exchange the quality of service (e.g., classification accuracy) for improvement in hardware parameters (e.g., latency or energy)~\cite{AxCinDNN:survey:2023}. 

Hardware accelerators of CNNs are implemented either as application-specific circuits (ASIC) or in field-programmable gate arrays (FPGA)~\cite{Sze:PEEE:17,Venkataramani:IEEEProc:2020,Capra:cnn:hw:survey:2020}. To reduce power consumption, hardware-oriented optimization techniques developed for CNNs focus on optimizing the data representation, pruning less important connections and neurons~\cite{pruning}, approximating arithmetic operations, compression of weights, and employing various smart data transfer and memory storage strategies~\cite{Sze:PEEE:17}. 

For example, according to~\cite{Sze:PEEE:17}, an 8-bit fixed-point multiply consumes 15.5$\times$ less energy (12.4$\times$ less area) than a 32-bit fixed-point multiply, and 18.5$\times$ less energy (27.5$\times$ less area) than a 32-bit floating-point multiply. Specialized number formats have been developed for training and inference~\cite{Venkataramani:IEEEProc:2020}. Further savings in energy are obtained not only by the bit width reduction of arithmetic operations but also by introducing approximate operations, particularly to the multiplication circuits~\cite{Sarwar:jestcs:18}. Note that the principles of approximate computing are usually introduced to existing CNN architecture after optimizing its accuracy. 

To integrate approximate components into NAS, EvoApproxNAS has been developed~\cite{Pinos:EvoApproxNAS}. It is a hardware-aware evolutionary NAS method employing Cartesian Genetic Programming (CGP) to evolve the CNN architectures for image classification tasks. Resulting CNNs are optimized in terms of architecture and can utilize approximate multipliers in convolutional layers. EvoApproxNAS employs NSGA-II to find the best trade-offs between the classification accuracy, the number of parameters (weights), and the inference energy. In particular, it considers the energy of multiplication because multiplication is the most energy-demanding arithmetic operation in CNNs. In addition to CNN architecture, it evolves the assignment of approximate multipliers to multiplication operations that are present in convolutional layers. The approximate multipliers are taken from the EvoApproxLib, which contains thousands of various implementations of approximate multipliers~\cite{EvoApproxLib}. On CIFAR-10 data set, EvoApproxNAS can deliver CNNs with approximate multiples, showing the same accuracy as a baseline implementation utilizing 8-bit exact multipliers but with energy reduced by $45\%$ (in the multiplication circuits) for one inference. EvoApproxNAS utilizes TFApprox4IL framework (see Section~\ref{sec:tfapprox}), which allows emulating the approximate multipliers (present in convolutional layers of CNNs) on common GPUs. 

\section{Proposed Approach}
\label{sec:proposed}

One of the main limitations of EvoApproxNAS is the search cost, which originates from the need to train (from scratch) and evaluate all candidate architectures during the search phase. Due to this, EvoApproxNAS is fairly limited in terms of the resulting CNN architecture size and complexity.
Another drawback of the EvoApproxNAS is its broad search space. Since this method searches for the whole CNN architectures (not the building blocks), the search space is too large and often contains nonsensical architectures.

In contrast, DARTS algorithm jointly learns architecture $\alpha$ and the weights $w$ (e.g., convolutional filter weights) and, therefore, drastically reduces the search cost compared to EvoApproxNAS. Additionally, DARTS search space is narrowed and mostly contains sensible architectures. As a consequence of that, DARTS is always able to produce some reasonable architecture quickly.

Before presenting the proposed DARTS methods with approximate multipliers, we briefly introduce the DARTS method and TFApprox4IL library.

\subsection{Preliminaries}

\subsubsection{DARTS method}
\label{sec:darts}

This work is built on the DARTS~\cite{DARTS} as the underlying NAS method for the automated design of CNNs. Following a proven approach, DARTS composes the resulting CNN architecture by stacking $L$ building blocks, called cells, sequentially one after the other.

A cell $C$ represents a directed acyclic graph (DAG) with $N$ nodes, defined as an ordered sequence $\{x^0, x^1, \cdots, x^{N-1}\}$, where each node $x^i$ is a latent representation. All directed edges $(i,j)$ are associated with some operation $o^{(i,j)}$, from a set of fixed operations $\mathcal{O}$ (e.g., zero, convolution or max pooling), that transforms $x^i$ to $x^j$.

Each cell $C_{i}$ has two input nodes, formed by the outputs of the two previous cells $C_{i-1}$ and $C_{i-2}$, and one output node, obtained by concatenating all intermediate nodes within a cell ${C_{i}}$. An intermediate node $x^{(j)}$ is computed as $$x^{(j)} = {\sum_{i<j}o^{(i,j)}(x^{(i)})}.$$

In order to make the search space continuous, DARTS introduces a set of continuous variables $\alpha = \{\alpha^{(i,j)}\}$ called architecture parameters. Each architecture parameter $\alpha^{(i,j)}$ is a vector of dimension $|\mathcal{O}|$ and serves as an operation mixing weight for a pair of nodes $x^i$ and $x^j$. The choice of an operation between nodes $x^i$ and $x^j$ is   computed as a softmax over all possible operations
$$
 \bar o^{(i,j)}(x) = \sum_{o \in \mathcal{O}}\frac{e^{\alpha_o^{(i,j)}}}{\sum_{o'\in \mathcal{O}}e^{\alpha_{o'}^{(i,j)}}}o(x).
$$

The search of the optimal cell is then reduced to learning the architecture parameters $\alpha$. This implies a bi-level optimization problem
\begin{equation*}
\begin{split}
& \min_{\alpha}\ \mathcal{L}_{val}(w^*(\alpha), \alpha) \\
& s.t.\  w^*(\alpha) = argmin_w\ \mathcal{L}_{train}(w,\alpha),
\end{split}
\end{equation*}
where $\mathcal{L}_{train}$, $\mathcal{L}_{val}$ denote the training and validation loss, respectively, and $w$ represents the weights associated with the architecture $\alpha$.
At the end of the search, the resulting cell architecture is obtained by keeping the most likely operations $o^{(i,j)}$ for each edge $(i,j)$.
\begin{figure*}[h]
    \centering
    \includegraphics[width=\textwidth]{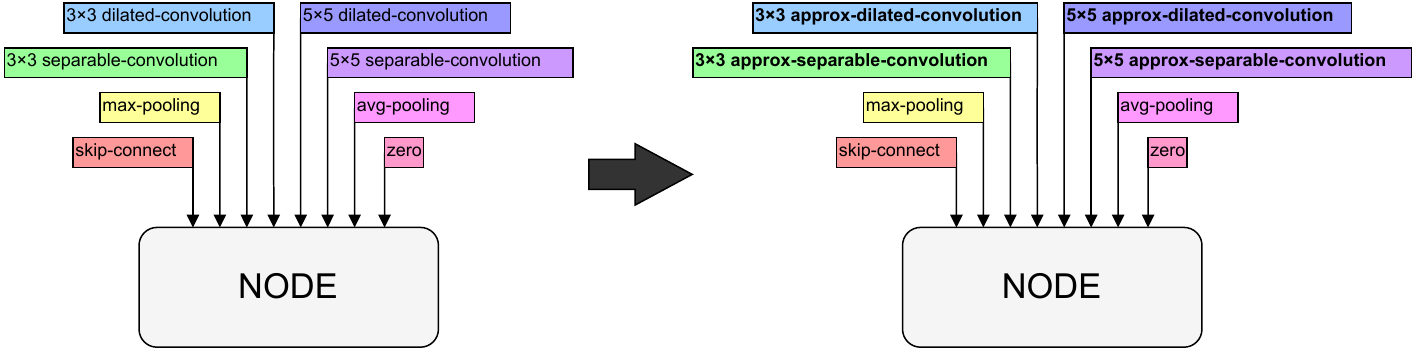}
    \caption{ApproxDARTS (right) replaces the $\{3\times3,\ 5\times5\}$ dilated convolutions and $\{3\times3,\ 5\times5\}$ separable convolutions of the original DARTS (left) with their approximate counterparts from the TFApprox4IL framework. All other operations (i.e., max pooling, avg pooling, zero, and skip connect) remain unchanged.}
    \label{fig:DARTS_ops}
\end{figure*}
\subsubsection{TFApprox4IL library}
\label{sec:tfapprox}

Due to the lack of native support for the approximate operations in modern DNN accelerators (e.g. GPUs), such as, for example, the employment of the approximate multiplications in convolutional layers of CNNs, a need for an emulator arises. For this purpose, a software library called TFApprox4IL~\cite{Pinos:tfapproxIL} was used.

TFApprox4IL supports the inference and training of  CNN models employing approximate multipliers in convolutional layers. At the core, TFApprox4IL emulates the approximate operations (i.e., multiplications) on GPU using LUT tables. To accelerate the computation of TFApprox4IL on GPUs, all kernels are highly optimized and written in CUDA. Additionally, the LUT tables are stored in the texture memory of the GPU to further improve the performance.

For an $8$-bit approximate multiplication, the computation flow of TFApprox4IL consists of multiple steps. First, the input floating-point operands $a$ and $b$ are converted to $8$-bit fixed point values $\hat{a}$ and $\hat{b}$, respectively, using the symmetric or asymmetric quantization scheme. Next, quantized operands $\hat{a}$ and $\hat{b}$ are combined to form a $16$-bit value, which serves as an address to a LUT table containing the result of approximate multiplication $\hat{a} \times \hat{b}$.

Natively, TFApprox4IL extends the Tensorflow\footnote{https://www.tensorflow.org/} Keras\footnote{https://keras.io} API and adds a support for \verb|ApproxConv2D| and \verb|ApproxDepthwiseConv2D| layers employing the approximate multipliers. However, more complex and convoluted layers can be composed by combining these layers (e.g., \verb|ApproxSeparableConv2D|).

\subsection{ApproxDARTS}

We propose a new extension of DARTS, called ApproxDARTS, which integrates the approximate multipliers into the original DARTS algorithm. For this purpose, we utilize the TFApprox4IL framework in order to employ the approximate multipliers into the architecture search stage of the DARTS. Following EvoApproxNAS, we use the approximate multipliers from the EvoApproxLibrary to substitute the accurate, 32-bit floating point multiplications in the convolutional layers with 8-bit fixed point approximate multiplications. The decision to substitute the multiplications only in the convolutional layers is based on the fact that the feature extraction part of the CNN models usually contains many multiplications. Due to this, a good trade-off between CNN accuracy and power consumption (of arithmetic operations) can be achieved. 

\begin{figure*}[ht]
    \centering
    \begin{subfigure}[t]{0.48\textwidth}
        \includegraphics[width=0.90\textwidth]{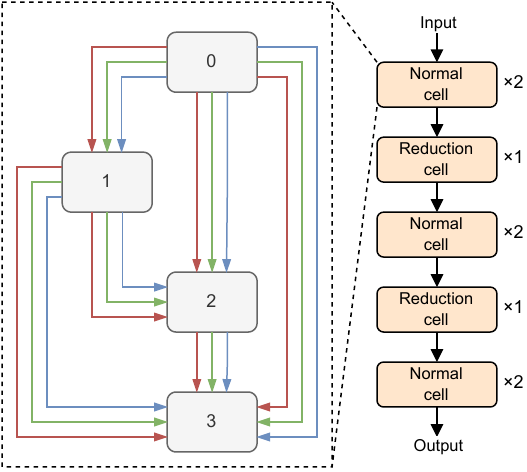}
        \caption{During the DARTS architecture search stage, a small model is constructed by stacking $8$ cells ($6$ normal cells and $2$ reduction cells). Each cell consists of $4$ nodes representing the latent features and $6$ sets of edges representing different operations (red, green, and blue) between the nodes. The DARTS method aims to find the best operation from the set of operations for every pair of nodes. }
        \label{fig:darts_search_stage}
    \end{subfigure}
    \hfill
    \begin{subfigure}[t]{0.48\textwidth}
        \includegraphics[width=0.90\textwidth]{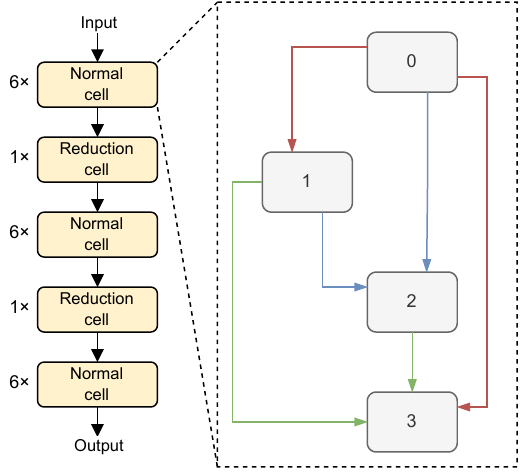}
        \caption{During the DARTS architecture evaluation stage, a final model is constructed by stacking $20$ cells ($18$ normal cells and $2$ reduction cells). Each cell consists of $4$ nodes representing the latent features, and $6$ edges, representing the operation (red, green or blue) between the nodes.}
        \label{fig:darts_eval_stage}
        \end{subfigure}

    \caption{The two stages of the DARTS method. 
    }
    \label{fig:darts}
\end{figure*}

 In accordance with the conventional NAS approaches and the original DARTS, a separate stage for the architecture search and the final architecture training is introduced. During the architecture search stage (Fig.~\ref{fig:darts_search_stage}), the training data set is evenly partitioned into two subsets, with one used for tuning the architecture and one for the optimization of the network weights (e.g., convolutional filter weights). The second stage is responsible for the final training of the optimal architecture, produced by the first stage, and utilizes the full training data set for training and the test data set for evaluation (Fig.~\ref{fig:darts_eval_stage}).


Four different approximate multipliers from the EvoApproxLib were selected. The selection of the approximate multipliers contains one accurate 8-bit multiplier \verb|mul8u_1JFF| (1JFF) as a baseline and three approximate 8-bit multipliers showing different trade-offs between the accuracy and power consumption, namely  \verb|mul8u_NGR| (NGR), \verb|mul8u_DM1| (DM1), and \verb|mul8u_2AC| (2AC). Selected approximate multipliers, together with their parameters, are listed in Table~\ref{tab:multipliers}.

\begin{table}[]
    \caption{Approximate 8-bit multipliers selected from the EvoApprox8b library. They are characterized in terms of the Mean Relative Error (MRE), Error Probability (EP), Mean Absolute Error (MAE), Worst Case Error (WCE), and energy consumption in a 45 nm technology at 1 V~\cite{EvoApproxLib}.}
    \centering
\begin{tblr}{
  row{1} = {c},
  cell{2}{2} = {c},
  cell{2}{3} = {c},
  cell{2}{4} = {c},
  cell{2}{5} = {c},
  cell{2}{6} = {c},
  cell{3}{2} = {c},
  cell{3}{3} = {c},
  cell{3}{4} = {c},
  cell{3}{5} = {c},
  cell{3}{6} = {c},
  cell{4}{2} = {c},
  cell{4}{3} = {c},
  cell{4}{4} = {c},
  cell{4}{5} = {c},
  cell{4}{6} = {c},
  cell{5}{2} = {c},
  cell{5}{3} = {c},
  cell{5}{4} = {c},
  cell{5}{5} = {c},
  cell{5}{6} = {c},
  hline{1,6} = {-}{0.08em},
  hline{2} = {-}{0.05em},
}
{\textbf{Multiplier}\\\textbf{}} & {\textbf{MRE}\\\textbf{[\%]}} & {\textbf{EP}\\\textbf{[\%]}} & {\textbf{MAE}\\\textbf{[\%]}} & {\textbf{WCE}\\\textbf{[\%]}} & {\textbf{Energy}\\\textbf{[mW]}} \\
mul8u\_1JFF                      & 0                             & 0                            & 0                             & 0                             & 0.391                            \\
mul8u\_2AC                       & 1.25                          & 98.12                        & 0.04                          & 0.12                          & 0.311                            \\
mul8u\_NGR                       & 1.90                          & 96.37                        & 0.07                          & 0.25                          & 0.276                            \\
mul8u\_DM1                       & 4.73                          & 98.16                        & 0.20                          & 0.89                          & 0.195                            
\end{tblr}
    \label{tab:multipliers}
\end{table}

ApproxDARTS uses the same set of operations $\mathcal{O}$ as the original DARTS implementation. This set consists of $8$ options, i.e. 
$\{3\times3,\ 5\times5\}$ \emph{separable convolution}, $\{3\times3,\ 5\times5\}$ \emph{dilated separable convolution}, $3\times3$ \emph{max pooling}, $3\times3$ \emph{average pooling}, \emph{skip connection}, and \emph{zero}. Using the TFApprox4IL framework, we reimplemented the separable convolution and dilated convolution to utilize the approximate multipliers in their kernels; see Fig.~\ref{fig:DARTS_ops}. For example, a separable convolution utilizing the approximate multipliers, \verb|ApproxSeparableConv2D|, can be implemented as \verb|ApproxDepthwiseConv2D| followed by $1\times1$ \verb|ApproxConv2D| operation from the TFApprox4IL.

Following this, we construct a set of approximate operations $\bar{\mathcal{O}}$ consisting of the same operations as the original set $\mathcal{O}$, except for the separable and dilated separable convolutions, which are substituted with their approximate counterparts. This set of approximate operations is used during both stages of the ApproxDARTS (i.e., the architecture search stage and the final architecture training stage).

During the architecture search stage of the ApproxDARTS method, a search for a suitable architecture of normal and reduction cells is conducted. In the evaluation stage, a large CNN is constructed by stacking several normal and reduction cells obtained from the architecture stage.

\section{Experimental Setup}
\label{sec:setup}

\subsection{Data sets}
Experiments are conducted on a popular image classification data set CIFAR-10~\cite{CIFAR-10}, consisting of $60\ 000$ RGB images with a spatial resolution of $32 \times 32$. The CIFAR-10 data set is split into the training subset consisting of $50\ 000$ images and the test subset with $10\ 000$ images. All images are evenly distributed over $10$ different classes.

\subsection{Hardware Platform}

All experiments were performed on a high-end GPU NVIDIA A100 with 40GB DRAM. For the architecture search stage of ApproxDARTS we used 2 GPUs, while for the subsequent architecture evaluation, we utilized 4 GPUs. The decision for this setup was made because it showed the best trade-off between the computation time and classification accuracy.  
While the parallelization highly improves the overall ApproxDARTS architecture search and evaluation time, it has a noticeable effect on the resulting CNN accuracy. This is due to the fact, that in order to fully utilize all available GPUs, the training batch size needs to be increased. However, too large batch sizes can cause lowered generalization of resulting CNNs~\cite{ImageNetInOneHour}. 

\section{Results}
\label{sec:results}

Results are presented for the architecture search and architecture evaluation separately in Sections~\ref{sec:approxdarts:search} and~\ref{sec:approxdarts:eval}. A comparison with EvoApproxNAS is discussed in  Section~\ref{sec:approxdarts:comp}. 

\subsection{Architecture search}
\label{sec:approxdarts:search}

A CNN model in the architecture search space is constructed by stacking $8$ cells ($6$ normal cells and $2$ reduction cells), where each cell consists of $N=6$ nodes. Reduction cells reduce the spatial resolution of feature maps by $2$, and are placed in $1/3$ and $2/3$ of the CNN architecture. The initial number of channels for this stage is set to $16$. 

We train this network for $50$ epochs with batch size $512$, during which the first $15$ epochs are used exclusively for the network weights (e.g., convolutional filters) optimization, while the remaining $35$ epochs jointly optimize the network weights and architecture parameters. A standard Stochastic Gradient Descent (SGD) optimizer with weight decay of $0.0003$ and momentum of $0.9$ is used for network weights optimization. The initial learning rate is $0.1$, which is annealed to $0$ following a cosine decay schedule. For optimizing architecture parameters $\alpha$, the Adam optimizer with a fixed learning rate of $0.0001$, the momentum of ($\beta_1=0.5$, $\beta_2=0.999$) and weight decay of $0.001$ is utilized.

\begin{figure*}
    \centering
    \begin{subfigure}[t]{0.45\textwidth}
        \includegraphics[width=0.95\textwidth]{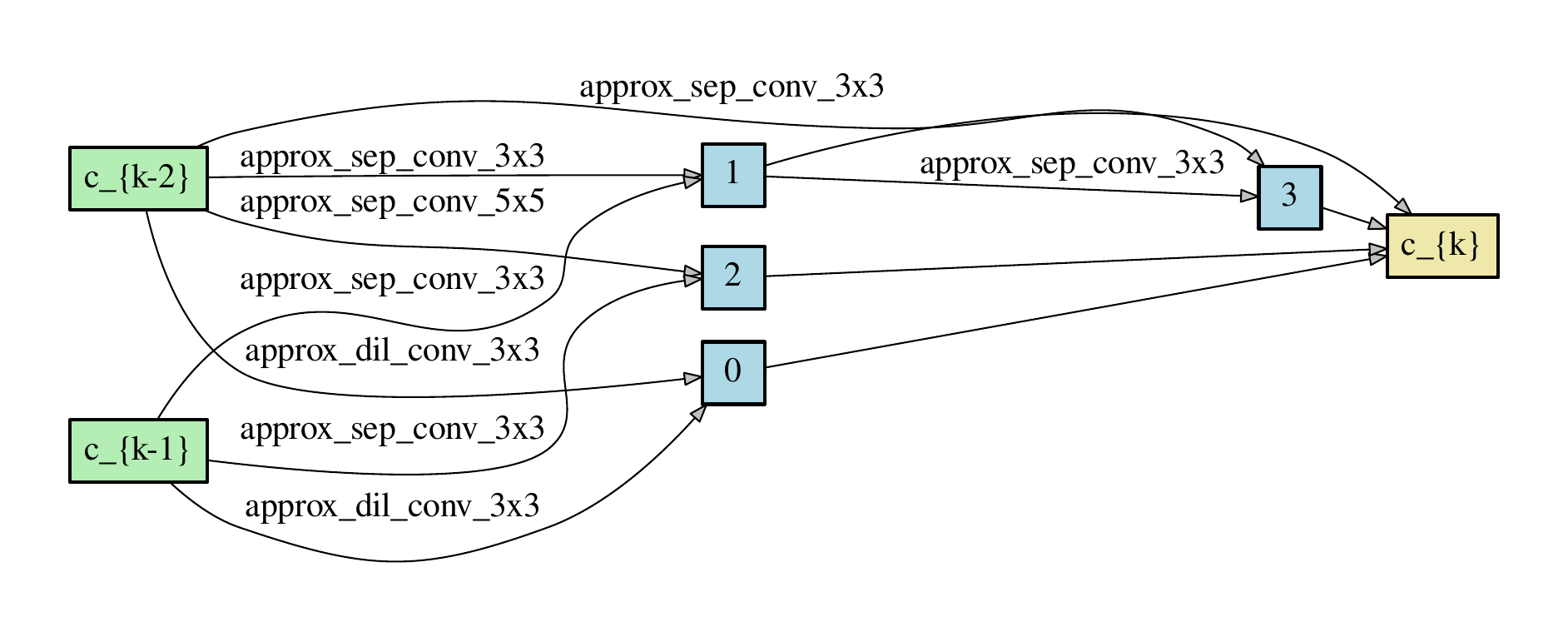}
        \caption{Best performing normal cell for the NGR multiplier.}
        \label{normal_cells_a}
    \end{subfigure}
    \begin{subfigure}[t]{0.45\textwidth}
        \includegraphics[width=0.95\textwidth]{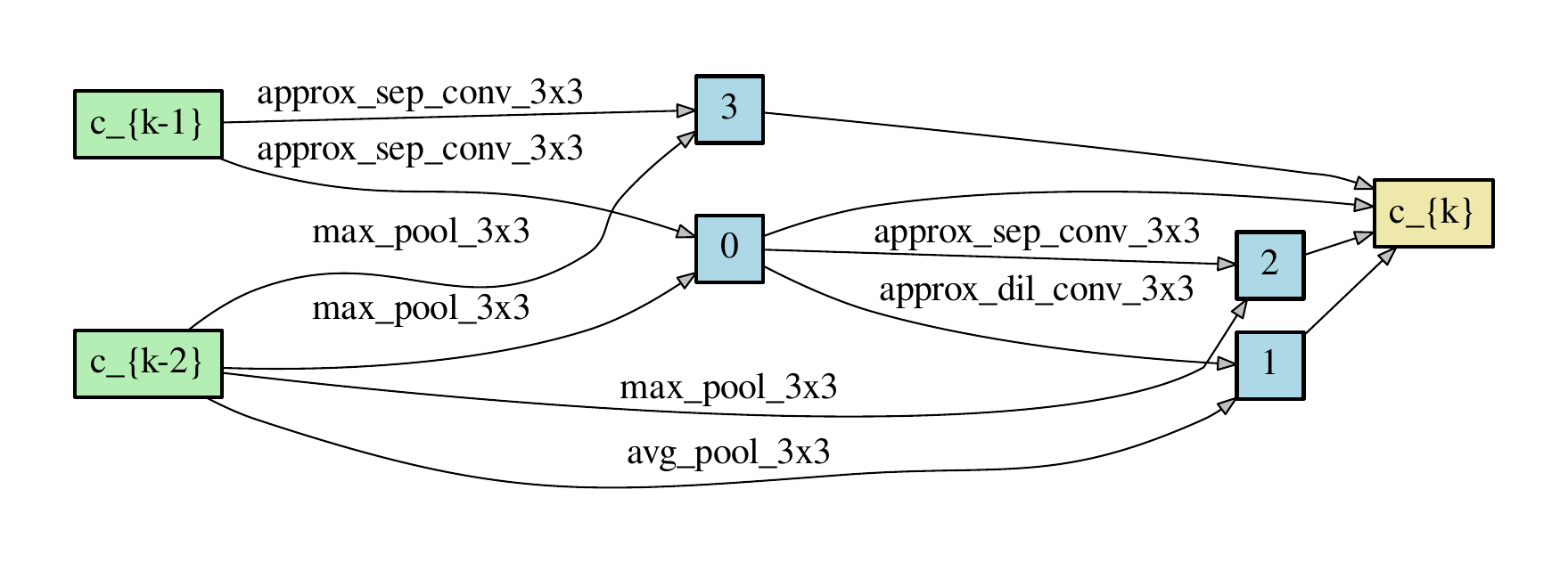}
        \caption{Best performing reduction cell for the NGR multiplier.}
        \label{reduction_cells_b}
        
    \end{subfigure}

    \begin{subfigure}[t]{0.45\textwidth}
        \includegraphics[width=0.95\textwidth]{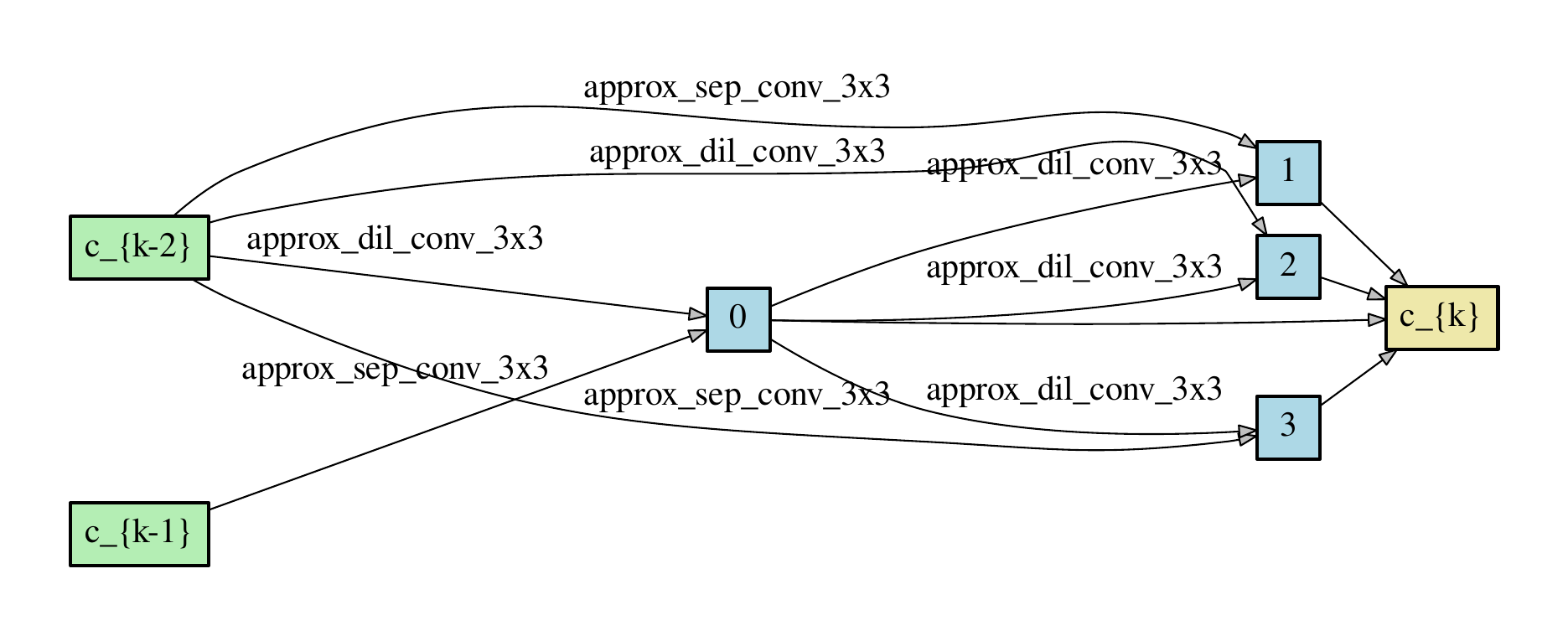}
        \caption{Best performing normal cell for the 2AC multiplier.}
        \label{normal_cells_c}
    \end{subfigure}
    \begin{subfigure}[t]{0.45\textwidth}
        \includegraphics[width=0.95\textwidth]{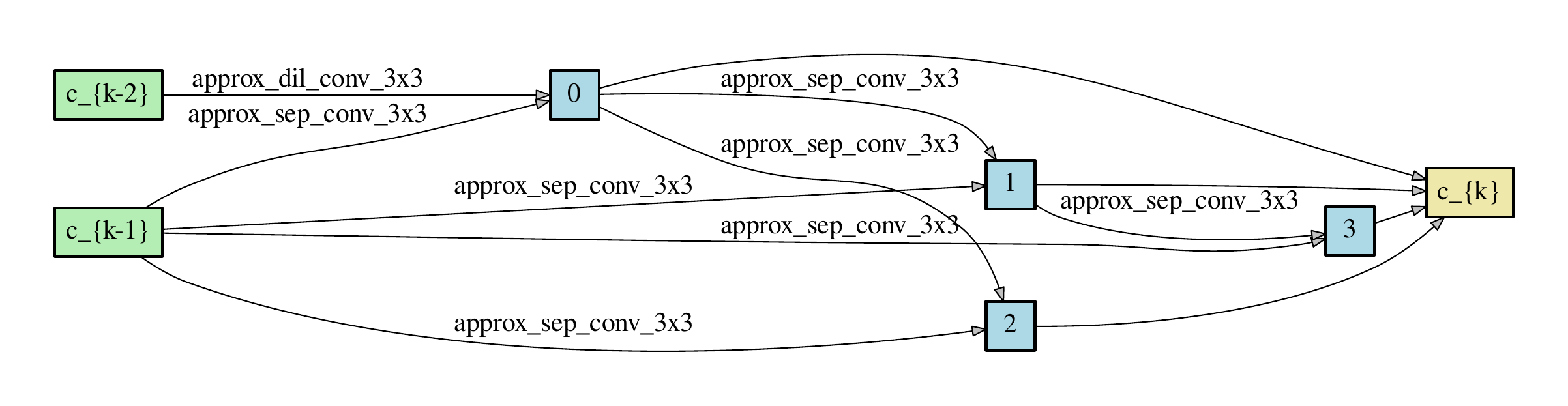}
        \caption{Best performing reduction cell for the 2AC multiplier.}
        \label{reduction_cells_d}
        
    \end{subfigure}
    
    \caption{Best performing  normal and reduction cells obtained during the architecture search stage of the ApproxDARTS for the NGR (\ref{normal_cells_a}, \ref{reduction_cells_b}) and 2AC (\ref{normal_cells_c}, \ref{reduction_cells_d}) approximate multipliers.}
    \label{fig:normal_cells}
\end{figure*}

A series of three runs of ApproxDARTS architecture search stage is performed for each of the selected approximate multipliers (i.e., \verb|mul8u_1JFF|, \verb|mul8u_NGR|, \verb|mul8u_DM1|, and \verb|mul8u_2AC|) separately. For comparison, we also perform three runs using the native, 32-bit floating point (\verb|FP32|) multiplication. Results of these experiments are summarized in Table~\ref{tab:runs_table}; see the Architecture Search Stage column for each run. The architecture search of the ApproxDARTS takes $7.5$ GPU hours on average for the \verb|FP32| multiplier, whereas the average time of architecture search is around $9.3$ GPU hours if the approximate multipliers are enabled. This slowdown is caused by the overhead of the TFApprox4IL framework. Figure~\ref{fig:normal_cells} shows the best-performing normal and reduction cells when NGR and 2AC multipliers are employed.

\begin{table}[t]
    \caption{ApproxDARTS method with various approximate multipliers: parameters of resulting CNNs on CIFAR-10 and GPU run-times.}
\resizebox{\columnwidth}{!}{
    \centering
\begin{tabular}{llclccc} 
\toprule[0.12em]
\multicolumn{1}{c}{\textbf{Run}} & \multicolumn{2}{c}{\textbf{Architecture Search Stage}} & \multicolumn{4}{c}{\textbf{Architecture Evaluation Stage}} \\ \cmidrule[0.1em](l){1-1} \cmidrule[0.1em](l){2-3} \cmidrule[0.1em](l){4-7}
\multicolumn{1}{c}{\begin{tabular}[c]{@{}c@{}}\textbf{\#}\\\textbf{}\end{tabular}} & \multicolumn{1}{c}{\begin{tabular}[c]{@{}c@{}}\textbf{Multiplier}\\\textbf{[-]}\end{tabular}} & \begin{tabular}[c]{@{}c@{}}\textbf{Cost}\\\textbf{[GPU hours]}\end{tabular} & \multicolumn{1}{c}{\begin{tabular}[c]{@{}c@{}}\textbf{Multiplier}\\\textbf{[-]}\end{tabular}} & \begin{tabular}[c]{@{}c@{}}\textbf{Params}\\\textbf{[M]}\end{tabular} & \begin{tabular}[c]{@{}c@{}}\textbf{Accuracy}\\\textbf{[\%]}\end{tabular} & \begin{tabular}[c]{@{}c@{}}\textbf{Cost}\\\textbf{[GPU days]}\end{tabular} \\\cmidrule[0.1em](l){1-1} \cmidrule[0.1em](l){2-2} \cmidrule[0.1em](l){3-3} \cmidrule[0.1em](l){4-4} \cmidrule[0.1em](l){5-5} \cmidrule[0.1em](l){6-6} \cmidrule[0.1em](l){7-7}
\multirow{9}{*}{1} & \multirow{5}{*}{FP32} & \multirow{5}{*}{7.55} & FP32 & \multirow{5}{*}{4.31} & $95.77$ & \multirow{5}{*}{2.81} \\
 &  &  & 1JFF &  & $95.24$ &  \\
 &  &  & NGR &  & $95.18$ &  \\
 &  &  & 2AC &  & $95.30$ &  \\
 &  &  & DM1 &  & $95.06$ &  \\ \cmidrule[0.06em](l){2-7}
 & 1JFF & $9.19$ & 1JFF & $4.77$ & $95.47$ & $2.28$ \\
 & NGR & $9.23$ & NGR & $4.72$ & $95.09$ & $2.14$ \\
 & 2AC & $9.21$ & 2AC & $4.61$ & $95.21$ & $2.24$ \\
 & DM1 & $9.33$ & DM1 & $5.02$ & $95.23$ & $2.26$ \\ 
\midrule[0.08em]
\multirow{9}{*}{2} & \multirow{5}{*}{FP32} & \multirow{5}{*}{7.58} & FP32 & \multirow{5}{*}{4.34} & $96.07$ & \multirow{5}{*}{2.97} \\
 &  &  & 1JFF&  & $95.50$ &  \\
 &  &  & NGR &  & $95.40$ &  \\
 &  &  & 2AC &  & $95.56$ &  \\
 &  &  & DM1 &  & $95.27$ &  \\\cmidrule[0.06em](l){2-7}
 & 1JFF& $9.18$ & 1JFF& $4.72$ & $95.71$ & $2.27$ \\
 & NGR & $9.25$ & NGR & $4.67$ & $95.82$ & $2.13$ \\
 & 2AC & $9.26$ & 2AC & $4.69$ & $95.33$ & $2.12$ \\
 & DM1 & $9.23$ & DM1 & $4.39$ & $95.07$ & $2.26$ \\ 
\midrule
\multirow{9}{*}{3} & \multirow{5}{*}{FP32} & \multirow{5}{*}{7.59} & FP32 & \multirow{5}{*}{4.10} & $96.05$ & \multirow{5}{*}{2.98} \\
 &  &  & 1JFF&  & $95.30$ &  \\
 &  &  & NGR &  & $95.32$ &  \\
 &  &  & 2AC &  & $95.37$ &  \\
 &  &  & DM1 &  & $95.30$ &  \\\cmidrule[0.06em](l){2-7}
 & 1JFF& $9.26$ & 1JFF& $5.01$ & $95.12$ & $2.11$ \\
 & NGR & $9.32$ & NGR & $5.05$ & $95.34$ & $1.95$ \\
 & 2AC & $9.23$ & 2AC & $4.30$ & $94.78$ & $2.32$ \\
 & DM1 & $9.10$ & DM1 & $4.73$ & $95.24$ & $2.19$ \\
\bottomrule[0.12em]
\end{tabular}
}
    \label{tab:runs_table}
\end{table}

\subsection{Architecture evaluation}
\label{sec:approxdarts:eval}

For the evaluation stage of ApproxDARTS, we construct a large CNN by stacking $20$ cells ($18$ normal cells and $2$ reduction cells) obtained during the architecture stage of ApproxDARTS. The reduction cells are placed in the $1/3$ and $2/3$ of the total depth of the CNN. Each cell consists of $N=6$ nodes. The initial number of channels for the model is $32$.

We train the network from scratch for $600$ epochs with a batch size of $256$. For the weights optimization, we utilize the SGD optimizer, with a weight decay of $0.003$ and a momentum of $0.9$. The initial learning rate is set to $0.025$ and is annealed down to $0$ following a cosine decay schedule. In order to prevent overfitting, we employ the $L2$ regularization~\cite{L2Regularization} for weights, drop path with probability $0.3$~\cite{devries2017}, cutout with size $16$~\cite{devries2017}, and auxiliary head with weight $0.4$.

For the best-performing architecture obtained during each of the architecture search stages, a series of nine evaluations is performed. In the case that the \verb|FP32| multiplier is used during the search stage, we evaluate the resulting architecture on all of the selected multipliers (i.e., \verb|mul8u_1JFF|, \verb|mul8u_NGR|, \verb|mul8u_DM1|, and \verb|mul8u_2AC|). In other cases, a single evaluation is employed using the multiplier utilized during the search stage. Table~\ref{tab:runs_table}, column \emph{Architecture Evaluation Stage} summarizes the evaluation stage's results. For each evaluation of architecture, the employed multiplier (\emph{Multiplier}), the number of parameters (\emph{Params}), the classification accuracy (\emph{Accuracy}), and the evaluation cost (\emph{Cost}), are reported. 

One can observe that the number of parameters is slightly higher when an approximate multiplier is utilized during the search. This implies, that the ApproxDARTS attempts to compensate the inaccuracies caused by the approximate multipliers by increasing the size of the architecture. Consequently, it can be noted that the classification accuracy drop caused by the employment of the approximate multipliers is low ($<1.3\%$). For the approximate multipliers, the highest classification accuracy of $95.82\%$ was reached for architecture designed with the \verb|mul8u_NGR| multiplier, whereas the lowest classification accuracy of $94.78\%$ was reported for architecture designed with the \verb|mul8u_2AC| multiplier. The overall highest classification accuracy of $96.07\%$ was achieved for the reference \verb|FP32| (exact) multiplier. 

\subsection{Comparison with EvoApproxNAS}
\label{sec:approxdarts:comp}

Table~\ref{tab:result_table} compares ApproxDARTS with EvoApproxNAS on CIFAR-10 data set. It has to be emphasized that a direct comparison is not possible because EvoApproxNAS is a multi-objective optimizer that can utilize various implementations of approximate multipliers during the search. ApproxDARTS always optimizes accuracy and uses one implementation for approximate multiplication. However, the following observations can be made:
\begin{itemize}
\item With only around $9.2$ GPU hours, ApproxDARTS requires less computational resources for the architecture search than EvoApproxNAS, which requires between $20.9 - 53.3$ GPU hours. 
\item When the accurate \verb|mul8u_1JFF| multiplier is utilized, ApproxDARTS achieves a higher classification accuracy ($95.43\%$) compared to the EvoApproxNAS's accuracy ($91.32\%$ when optimized for Accuracy and Energy; and $91.54\%$ when optimized for Accuracy, Energy, and Parameters). 
\item In the case of the approximate multipliers, ApproxDARTS achieves the highest classification accuracy of $95.42\%$ for the \verb|mul8u_NGR| multiplier, $95.11\%$ for the \verb|mul8u_2AC| multiplier, and $95.18\%$ for the \verb|mul8u_DM1| multiplier, which outperforms the EvoApproxNAS highest achieved classification accuracy of $91.14\%$ and $87.55\%$ (depending on the optimization objectives; see Table~\ref{tab:result_table}). 
\item By examining the size of the architectures in Table~\ref{tab:result_table}, it can be observed that the ApproxDARTS creates larger networks, consisting of $4.25$M - $4.83$M parameters, whereas the number of parameters of CNNs generated by EvoApproxNAS is ranging from $0.35$M to $2.07$M. The increase in the size of CNNs designed by ApproxDARTS originates from the fact that ApproxDARTS is a single objective method (leaving thus the CNN size unoptimized), and we always considered the default (and probably unnecessarily large) number of cells as in the original DARTS implementation. 
\end{itemize}

\begin{table}[]
    \caption{A comparison of the results of ApproxDARTS and EvoApproxNAS on the CIFAR-10 data set.}
    \centering
    \resizebox{0.5\textwidth}{!}{
        \centering
    
\begin{tblr}{
  row{1} = {c},
  cell{2}{2} = {c},
  cell{2}{4} = {c},
  cell{2}{5} = {c},
  cell{2}{6} = {c},
  cell{3}{2} = {c},
  cell{3}{4} = {c},
  cell{3}{5} = {c},
  cell{3}{6} = {c},
  cell{4}{2} = {c},
  cell{4}{4} = {c},
  cell{4}{5} = {c},
  cell{4}{6} = {c},
  cell{5}{2} = {c},
  cell{5}{4} = {c},
  cell{5}{5} = {c},
  cell{5}{6} = {c},
  cell{6}{2} = {c},
  cell{6}{4} = {c},
  cell{6}{5} = {c},
  cell{6}{6} = {c},
  cell{7}{2} = {c},
  cell{7}{4} = {c},
  cell{7}{5} = {c},
  cell{7}{6} = {c},
  cell{8}{2} = {c},
  cell{8}{4} = {c},
  cell{8}{5} = {c},
  cell{8}{6} = {c},
  cell{9}{2} = {c},
  cell{9}{4} = {c},
  cell{9}{5} = {c},
  cell{9}{6} = {c},
  cell{10}{2} = {c},
  cell{10}{4} = {c},
  cell{10}{5} = {c},
  cell{10}{6} = {c},
  cell{11}{1} = {c=6}{},
  hline{1,11} = {-}{0.08em},
  hline{2,7,9} = {-}{0.05em},
}
{\textbf{Method}\\\textbf{}} & {\textbf{Objectives}\footnote[2]{}\\\textbf{}} & {\textbf{Multiplier}\\\textbf{}} & {\textbf{Accuracy}\\\textbf{(\%)}} & {\textbf{Params}\\\textbf{(M)}} & {\textbf{Search Cost}\\\textbf{(GPU hours)}} \\
ApproxDARTS                  & A                          & FP32                   & $95.96\pm0.14$                               & $4.25$                      & $7.57$                                  \\
ApproxDARTS                  & A                          & 1JFF                      & $95.43\pm0.09$                               & $4.83$                      & $9.21$                                  \\
ApproxDARTS                  & A                          & NGR                       & $95.42\pm0.13$                               & $4.81$                      & $9.27$                                   \\
ApproxDARTS                  & A                          & 2AC                       & $95.11\pm0.14$                               & $4.53$                      & $9.23$                                  \\
ApproxDARTS                  & A                          & DM1                       & $95.18\pm0.03$                               & $4.71$                      & $9.22$                                   \\
EvoApproxNAS                 & A/E                        & 1JFF                      & 91.32                              & 0.84                            & $26.8$                                       \\
EvoApproxNAS                 & A/E                        & multiple                         & 91.14                              & 2.07                            & $53.3$                                       \\
EvoApproxNAS                 & A/E/P                      & 1JFF                      & 91.54                              & 1.33                            & $36.3$                                       \\
EvoApproxNAS                 & A/E/P                      & multiple                         & 87.55                              & 0.35                            & $20.9$                             \\
{\scriptsize \footnote[2]{} Objectives are classification accuracy (A), number of parameters (P) and energy (E).}
\end{tblr}
    }
    \label{tab:result_table}
\end{table}


For example, the best-performing architecture delivered by the ApproxDARTS contains $4.67$M parameters and achieves the accuracy of $95.82\%$ when the \verb|mul8u_NGR| approximate multiplier is employed, and $96.22\%$ for the \verb|FP32| multiplier. Using the TensorFlow Profiler we analyzed the model and revealed that it must perform $1.18$B FLOPs in one inference, from which $706.2$M FLOPS are spent in the approximate layers, and $471.57$M FLOPS are additional computations (e.g., the batch normalizations, $1\times1$ convolutions, etc.). Therefore, the TFApprox4IL approximates around $60\%$ of operations performed by this architecture; the rest is computed using the native $32$-bit floating point (\verb|FP32|) arithmetic. By employing the \verb|mul8u_NGR| approximate multiplier instead of the \verb|FP32| exact multiplier in the convolutional layers, a CNN delivered by ApproxDARTS can save $53.84$\% of energy in arithmetic operations. Using the \verb|mul8u_1JFF| instead of the \verb|mul8u_NGR|, the energy saving would be $50.90\%$. 

The highest-accurate CNN delivered by EvoApproxNAS with approximate multipliers shows $91.14\%$ accuracy, $2.07$M parameters, and energy consumption of $220.9~\mu$J (in arithmetic operations) for one inference~\cite{Pinos:EvoApproxNAS}. The best CNN utilizing approximate multiplication created by ApproxDARTS shows the $95.82\%$ accuracy and $4.67$M parameters; its inference consumes $267.03~\mu$J in terms of arithmetic operations\footnote{We used TensorFlow Profiler tools to count all FLOPS associated with the approximate layers.}.

\section{Conclusions}
\label{sec:conclusions}

In this work, we presented ApproxDARTS, a neural network architecture search algorithm combining the popular DARTS method with the deployment of approximate multipliers enabled by the TFApprox4IL framework. 
We showed on the CIFAR-10 data set that the ApproxDARTS is able to perform a complete architecture search within less than $10$ GPU hours and produce CNN architectures with similar complexity (and a small classification accuracy drop of $<1.3\%$) compared to our reference implementation of the original method. Additionally, by employing approximate multipliers, the ApproxDARTS method can achieve (i) a significant energy consumption reduction of $53.84\%$ in the arithmetic operations of the inference phase of the resulting CNN, compared to the utilization of the native $32$-bit floating point multiplier and (ii) an energy consumption reduction of $5.97\%$ compared to the utilization of the exact $8$-bit fixed point multiplier; in both cases with a negligible accuracy drop.

As the proposed method explicitly optimized neither CNN parameters nor energy, we plan to integrate the network complexity and energy into the optimization process in our future work. One approach is to constrain the number of cells; another is modifying the cost function of DARTS. Furthermore, we will evaluate ApproxDARTS with more challenging data sets.

\section*{Acknowledgment}
This work was supported by Czech Science Foundation project GA24-10990S and Ministry of Education, Youth and Sports of Czech Republic through the e-INFRA CZ (ID:90254).



%

\bibliographystyle{IEEEtr}
\bibliography{paper}

\end{document}